\documentclass[AMA,Times1COL]{WileyNJDv5} 

\articletype{Review Article}%

\received{Date Month Year}
\revised{Date Month Year}
\accepted{Date Month Year}
\journal{}
\volume{}
\copyyear{2024}
\startpage{1}

\begin{document}

\title{GPU Based Differential Evolution: New Insights and Comparative Study}

\author[1]{Dylan M. Janssen}

\author[2,3]{Wayne Pullan}

\author[3]{Alan Wee-Chung Liew}

\authormark{JANSSEN \textsc{et al.}}
\titlemark{GPU Based Differential Evolution: New Insights and Comparative Study}

\address{\orgdiv{School of Information and Communication Technology}, \orgname{Griffith University}, \orgaddress{\state{Queensland}, \country{Australia}}}

\corres{Corresponding author Alan Wee-Chung Liew. \email{a.liew@griffith.edu.au}}

\abstract[Abstract]
{
    Differential Evolution (DE) is a highly successful population based global optimisation algorithm, commonly used for solving numerical optimisation problems. However, as the complexity of the objective function increases, the wall-clock run-time of the algorithm suffers as many fitness function evaluations must take place to effectively explore the search space. Due to the inherently parallel nature of the DE algorithm, graphics processing units (GPU) have been used to effectively accelerate both the fitness evaluation and DE algorithm. This work reviews the main architectural choices made in the literature for GPU based DE algorithms and introduces a new GPU based numerical optimisation benchmark to evaluate and compare GPU based DE algorithms. 
}

\keywords{Differential Evolution, Graphics Processing Units, CUDA, Numerical Optimisation}

\jnlcitation{\cname{%
\author{Janssen D.},
\author{Pullan W.}, and
\author{Liew A.W-C.}}.
\ctitle{GPU Based Differential Evolution: New Insights and Comparative Study.} \cjournal{\it J Concurrency Computat Pract Exper.} \cvol{2023;00(00):1--18}.}

\maketitle

\renewcommand\thefootnote{}
\renewcommand\thefootnote{\fnsymbol{footnote}}
\setcounter{footnote}{1}

\section{Introduction}\label{sec1}

Differential evolution (DE) is a stochastic evolutionary optimisation algorithm that can be used to solve challenging optimisation problems. DE maintains a population of solutions and uses specific mutation, crossover, and selection operators to guide the population toward near-optimal solutions. Since DE is a black box optimisation algorithm, recent studies have focused on visualising the convergence properties of DE, to gain a better understanding of how DE behaves throughout the optimisation process \cite{ref:2023:janssen}. 
 
\begin{figure*}[!t]
    \centering
    \includegraphics[width=\textwidth]{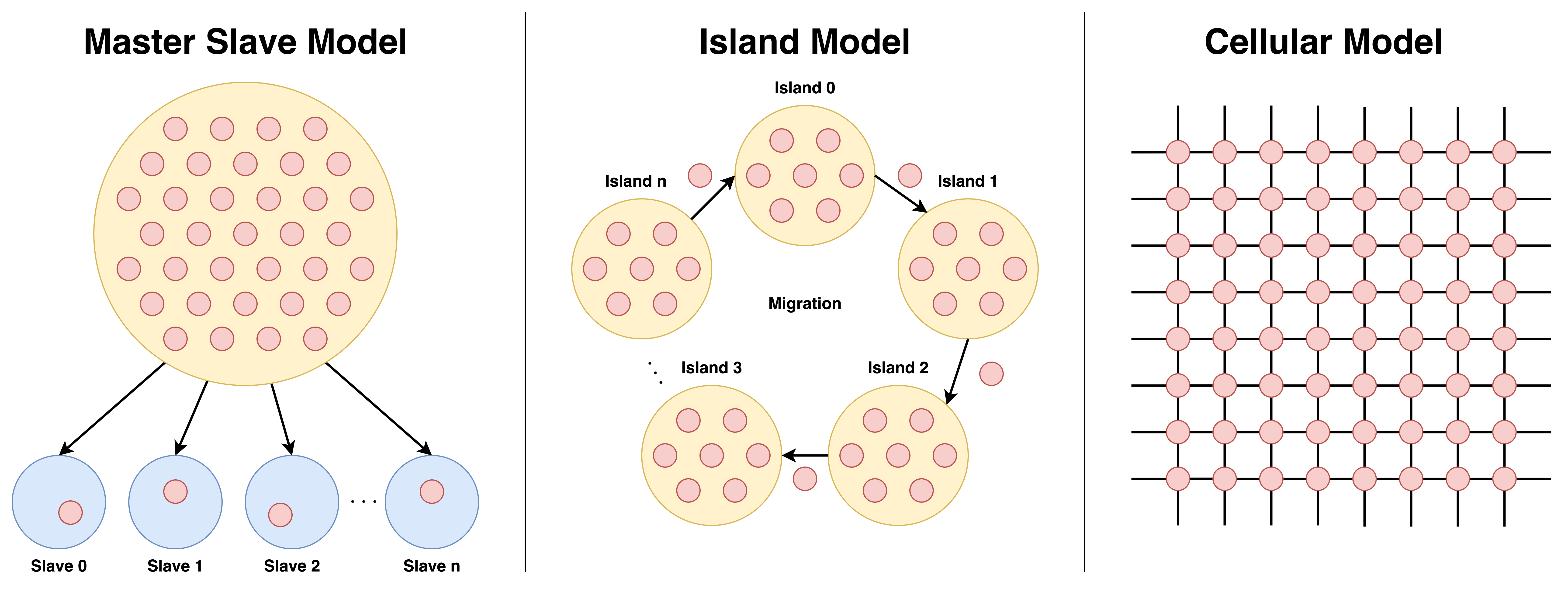}
    \caption{Parallel models for Evolutionary Algorithms}
    \label{fig:Parallel-Genetic-Algorithm-Types}
\end{figure*}

Although DE has proven to be highly successful in solving optimisation problems, the population-based approach is inherently slow as many solutions have to be evaluated and can result in lengthy execution times. One method of addressing this performance issue is to take advantage of the fact that evolutionary algorithms can be easily adapted to a parallel framework and distribute the computation over a number of computing units. This approach has been used for CPU based EAs and three canonical parallel models that have evolved : Master Slave; Island Model; and Cellular Model\cite{ref:2000:cantu}, which are depicted in Fig. \ref{fig:Parallel-Genetic-Algorithm-Types}. The master slave model is very similar to a regular EA, where the main improvement is a wall clock time speedup. However, the island and cellular parallel models introduce new concepts that influence convergence properties as well as providing a wall-clock speedup. These parallel models are distinguished by their parallel grain - the relationship between computation and communication.

With the advent of Graphics Processing Units (GPUs), the opportunity has arisen for the computation to be cheaply spread over many thousand processing units. In DE, this GPU approach has been utilised over the last 10 years with a major focus on designing algorithms that can effectively use the Single Instruction Multiple Thread (SIMT) parallel architecture of Nvidia GPUs. 

This study carries out a comparison of select GPU based DE algorithms, implemented in the period 2010 - 2023 time-frame, with specific reference to the choices made to map the DE algorithm to the Nvidia GPU architecture (the last review of GPU accelerated DE was conducted in 2013 \cite{ref:2013:kromer}). In addition, this study evaluates how different GPU based DEs have been compared and proposes a new method for performing these evaluations. 

This paper is structured as follows: 
Section 2 is a background of DE algorithms while 
Section 3 presents an overview of parallel models for EAs
Section 4 provides an overview of Nvidia GPU hardware and software architecture. 
Section 5 describes parallel models for GPU based evolutonary algorithms and 
Section 6 discusses existing GPU based DE algorithms and 
Section 7 presents the different ways these GPU based DE algorithms have been benchmarked and 
introduces a new GPU based numerical optimisation benchmark with 
Section 8 presenting case studies using the proposed numerical optimisation benchmark. 
Finally Section 9 provides a conclusion and suggestions for future work in GPU based DE. 

\section{Differential Evolution Algorithms}\label{sec2}

In 1997 a simple but powerful population-based search algorithm called DE was proposed \cite{ref:1997:storn}. DE consists of $N_P$ individuals, where each individual contains a $D$ dimensional real valued parameter vector representing a possible solution to a given problem. 

Firstly, the population is randomly initialised within the bounds of the search space and each individuals objective value is evaluated. Each individual or \textit{target vector} then undergoes three bio-inspired operations called mutation, crossover, and selection, in an evolutionary loop until some termination criteria is met. 

\begin{itemize}
    \item \textbf{Mutation}: a \textit{base vector} is selected from the population as the initial reference point for the mutation and the difference (or \textit{differential}) of randomly selected population members excluding the \textit{target vector} are scaled by a constant $F$ and added to the \textit{base vector}, producing a \textit{mutant vector}.
    \item \textbf{Crossover}: applied between the \textit{target vector} and \textit{mutant vector} with a probability $C_R$ to generate a \textit{trial vector}. 
    \item \textbf{Selection}: the \textit{trial vector} is evaluated and replaces the \textit{target vector} in the next generation if its objective value is higher quality. Otherwise, the \textit{target vector} remains in the population. 
\end{itemize}

Numerous mutation and crossover strategies have been proposed and a classification scheme has been developed using the notation: 
\begin{equation}
    DE/x/y/z
\end{equation}
where $x$ is the vector to be mutated, usually \textit{rand} (randomly selected individual) or \textit{best} (best objective value individual); $y$ is the number of difference vectors used in the mutation; and $z$ is the crossover scheme, e.g. \textit{bin} for binomial crossover, or \textit{exp} for exponential crossover. 

The most popular strategy is $DE/rand/1/bin$ where a randomly selected population member is used as the base vector and is scaled by the vector difference of two randomly selected individuals to generate a mutant vector: 

\begin{equation}
    y_i = x_{r_1} + F(x_{r_2} - x_{r_3})
\label{eq:rand_mutation}
\end{equation}

where $x_{r_1}$ $x_{r_2}$ $x_{r_3}$ are three mutually exclusive random individuals that also differ from the target vector, $F$ is the differential weight factor, and $y_i$ is the mutant vector. Binomial crossover is then performed to create the \textit{trial vector}: 

\begin{equation}
    z_{i,j} = 
    \begin{cases}
        y_{i,j} & \text{if } \text{rand}_{j} \leq C_R \text{ or } j=j_{\text{rand}} \\
        x_{i,j} & \text{otherwise}
    \end{cases}
\label{eq:crossover}
\end{equation}

where $i$ represents the index of the vector in the population and $j$ is the position in the vector, $\text{rand}_{j}$ is a random real number drawn from a uniform distribution in the range $[0, 1]$, $j_{\text{rand}}$ is a random integer drawn from a uniform distribution in the range $[0, D)$ and $C_R$ controls the number of parameters that are transferred during the crossover operation. 

Many DE variations have been proposed such as:
\begin{itemize}
    \item \textbf{Self-adaptive DE (SaDE)}: learning strategy and control parameters are gradually self-adapted at a population level \cite{ref:2005:qin};
    \item \textbf{jDE}: control parameters are adapted at an individual level \cite{ref:2006:brest}; 
    \item \textbf{Chaotic DE}: uses the properties of a chaotic system to spread individuals throughout the search space \cite{ref:2007:wang}; 
    \item \textbf{JADE}: introduces a new mutation strategy, adapts the control parameters and uses an optional archive of previous individuals to enhance search diversity \cite{ref:2009:zhang};

\end{itemize}

\section{Parallel models for Evolutionary Algorithms}
Before a review of GPU based DE is presented, the general models that have been used to implement parallel evolutionary algorithms (EA) will be defined. This will allow a clearer distinction of where GPU based DE research fits within the larger field of parallel EAs. Prior to the advent of general purpose computing on GPUs, progress in parallel EAs has been made using compute clusters with many networked nodes and modern multi-core CPUs. There are three canonical parallel models: Master Slave; Island Model; and Cellular Model\cite{ref:2000:cantu}, which are depicted in Fig. \ref{fig:Parallel-Genetic-Algorithm-Types}. The master slave model is very similar to a regular EA, where the main improvement is a wall clock time speedup. However, the island and cellular parallel models introduce new concepts that influence convergence properties as well as providing a wall-clock speedup. These parallel models are distinguished by their parallel grain - the relationship between computation and communication.

\subsection{Master Slave model}
The master slave model for parallel EAs maintains a single population, where one process manages the population and dispatches work to slave processes. Typically, as the fitness evaluation is the most computationally demanding aspect of an EA it is dispatched to the slave processes, however sometimes genetic operators are applied in parallel as well. The master slave model works well for problems with complex fitness functions and a relatively small number of slave processes as the master process can become a bottleneck if too much communication is required. Synchronous and Asynchronous approaches can be used for this model and have different characteristics. The synchronous approach is when the master process waits for all evaluations to complete before continuing to the next generation, mimicking a regular EA, where a wall-clock speedup is the major difference. The asynchronous approach however removes the synchronisation and instead uses available data, without waiting for all slave processes to finish before moving to the next generation. The asynchronous master slave model has less communication overhead, however is almost impossible to duplicate results due to the scheduling of the slave processes.

\subsection{Island Model} 
The island model, also known as the \textit{coarse grained} model, consists of decentralising a larger population by partitioning it into several sub-populations that evolve independently and are interconnected to periodically share individuals \cite{ref:1995:belding}. There are many topologies for the interconnections including random\cite{ref:2004:tang}, ring\cite{ref:1998:paz}, roulette wheel\cite{ref:2013:lopes}, tournament\cite{ref:2013:lopes}, and more. Additional parameters are introduced controlling when migration occurs and how migrants are selected and synchronous and asynchronous migration strategies exist. The island model was developed for a networked cluster of compute nodes due to the infrequent communication required for the migration operator but can also be implemented on modern multi-core CPUs. The isolated sub-populations with periodic migrations provide a good balance between exploration and exploitation as the separate sub-population searches occur in several regions of the search space.

\subsection{Cellular Model}
The cellular model, also known as the \textit{fine grained} model, uses a spatial arrangement of overlapped small neighbourhoods to aid in exploring the search space \cite{ref:1998:paz}. Each individual is provided its own pool of potential mates defined by the neighbouring individuals, while simultaneously each individual belongs to many pools. Commonly, the population is arranged in a one-dimensional or two-dimensional structure with overlapped neighbourhoods to provide a smooth diffusion of solutions across the population. Each individual is processed in parallel, where each individual can only interact with its direct neighbours, and the replacement operator overwrites the considered individual. This model is suited for distributed memory MIMD \cite{ref:1993:maruyama} and SIMD computers due to the high-level of communication required. The local selective pressure and the individuals being isolated by distance provides the cellular model with a highly-diverse population.

\section{Nvidia GPU Hardware and Software}

\begin{table*}[!t]
\caption{Summary of Selected Differences Between Nvidia GPU Compute Capabilities.}
\label{tab:compute_capability}
\centering 
\resizebox{\textwidth}{!}
{
    \begin{tabular}{|l|ccccccccccccccccccccccc|}
    \hline
    \multicolumn{1}{|c|}{\multirow{2}{*}{\textbf{Technical specifications}}} &
      \multicolumn{23}{c|}{\textbf{Compute capability (version)}} \\ \cline{2-24} 
    \multicolumn{1}{|c|}{} &
      \multicolumn{1}{c|}{\textbf{1}} &
      \multicolumn{1}{c|}{\textbf{1.1}} &
      \multicolumn{1}{c|}{\textbf{1.2}} &
      \multicolumn{1}{c|}{\textbf{1.3}} &
      \multicolumn{1}{c|}{\textbf{2.x}} &
      \multicolumn{1}{c|}{\textbf{3}} &
      \multicolumn{1}{c|}{\textbf{3.2}} &
      \multicolumn{1}{c|}{\textbf{3.5}} &
      \multicolumn{1}{c|}{\textbf{3.7}} &
      \multicolumn{1}{c|}{\textbf{5}} &
      \multicolumn{1}{c|}{\textbf{5.2}} &
      \multicolumn{1}{c|}{\textbf{5.3}} &
      \multicolumn{1}{c|}{\textbf{6}} &
      \multicolumn{1}{c|}{\textbf{6.1}} &
      \multicolumn{1}{c|}{\textbf{6.2}} &
      \multicolumn{1}{c|}{\textbf{7}} &
      \multicolumn{1}{c|}{\textbf{7.2}} &
      \multicolumn{1}{c|}{\textbf{7.5}} &
      \multicolumn{1}{c|}{\textbf{8}} &
      \multicolumn{1}{c|}{\textbf{8.6}} &
      \multicolumn{1}{c|}{\textbf{8.7}} &
      \multicolumn{1}{c|}{\textbf{8.9}} &
      \textbf{9} \\ \hline
    \begin{tabular}[c]{@{}l@{}}Maximum number \\ of threads per block\end{tabular} &
      \multicolumn{4}{c|}{512} &
      \multicolumn{19}{c|}{1024} \\ \hline
    \begin{tabular}[c]{@{}l@{}}Maximum number \\ of resident blocks \\ per multiprocessor\end{tabular} &
      \multicolumn{5}{c|}{8} &
      \multicolumn{4}{c|}{16} &
      \multicolumn{8}{c|}{32} &
      \multicolumn{1}{c|}{16} &
      \multicolumn{1}{c|}{32} &
      \multicolumn{2}{c|}{16} &
      \multicolumn{1}{c|}{24} &
      32 \\ \hline
    \begin{tabular}[c]{@{}l@{}}Maximum number \\ of 32-bit registers \\ per thread block\end{tabular} &
      \multicolumn{2}{c|}{8 K} &
      \multicolumn{2}{c|}{16 K} &
      \multicolumn{1}{c|}{32 K} &
      \multicolumn{1}{c|}{64 K} &
      \multicolumn{1}{c|}{32 K} &
      \multicolumn{4}{c|}{64 K} &
      \multicolumn{1}{c|}{32 K} &
      \multicolumn{2}{c|}{64 K} &
      \multicolumn{1}{c|}{32 K} &
      \multicolumn{8}{c|}{64 K} \\ \hline
    \begin{tabular}[c]{@{}l@{}}Maximum number \\ of 32-bit registers \\ per thread\end{tabular} &
      \multicolumn{4}{c|}{124} &
      \multicolumn{2}{c|}{63} &
      \multicolumn{17}{c|}{255} \\ \hline
    \begin{tabular}[c]{@{}l@{}}Maximum amount \\ of shared memory \\ per thread block\end{tabular} &
      \multicolumn{4}{c|}{16 KB} &
      \multicolumn{11}{c|}{48 KB} &
      \multicolumn{1}{c|}{96 KB} &
      \multicolumn{1}{c|}{48 KB} &
      \multicolumn{1}{c|}{64 KB} &
      \multicolumn{1}{c|}{163 KB} &
      \multicolumn{1}{c|}{99 KB} &
      \multicolumn{1}{c|}{163 KB} &
      \multicolumn{1}{c|}{99 KB} &
      227 KB \\ \hline
    \begin{tabular}[c]{@{}l@{}}Number of shared \\ memory banks\end{tabular} &
      \multicolumn{4}{c|}{16} &
      \multicolumn{19}{c|}{32} \\ \hline
    \begin{tabular}[c]{@{}l@{}}Amount of local \\ memory per thread\end{tabular} &
      \multicolumn{4}{c|}{16 KB} &
      \multicolumn{19}{c|}{512 KB} \\ \hline
    Constant memory size &
      \multicolumn{23}{c|}{64 KB} \\ \hline
    \end{tabular}%
}
\end{table*}

CUDA is a parallel computing platform and programming model developed by NVIDIA for general computing on GPUs. Based on the widely used C/C++ and Fortran programming languages, CUDA enables programs to create and run functions (referred to as "kernels") for general-purpose computation on Nvidia GPUs. Note that, in CUDA, the term "host" refers to the CPU and its memory, while the term "device" refers to the GPU and its memory. 

Code run on the host can manage memory on both the host and device, and also launch kernels to be executed on the device. These kernels are executed by many GPU threads in parallel where threads are organised into thread blocks within which they are grouped into warps. The most important consideration for algorithm development on GPUs is instructions are issued to warps and executed one common instruction at a time in lockstep (SIMD), meaning that the threads of a warp should have minimal divergence or branching. 

The processing flow of a CUDA program is as follows: 
\begin{itemize}
\item copy the required data over the high-speed PCI-E bus (see Fig. \ref{fig:gpu_mem_levels}) from the host to the device; 
\item launch compute kernels on the device; 
\item copy the output of the kernel(s) to the host over PCI-E; 
\item and finally reset the device. 
\end{itemize}

To effectively implement a CUDA program careful attention must be made to the design of the kernels as: 
\begin{itemize}
\item data transfer between host and device is limited by the bandwidth of PCI-E and should be minimised for maximum performance; 
\item the number of threads should be maximised; 
\item threads in a warp should have minimum divergence; 
\item global memory accesses should be coalesced; 
\item shared memory use should be maximised; and
\item shared memory bank conflicts should be avoided. 
\end{itemize}
The following sub-sections outline the Nvidia GPU architecture, thread hierarchy, memory hierarchy, and random number generation. 

\begin{figure}[!t]
    \centering
    \includegraphics[width=\textwidth]{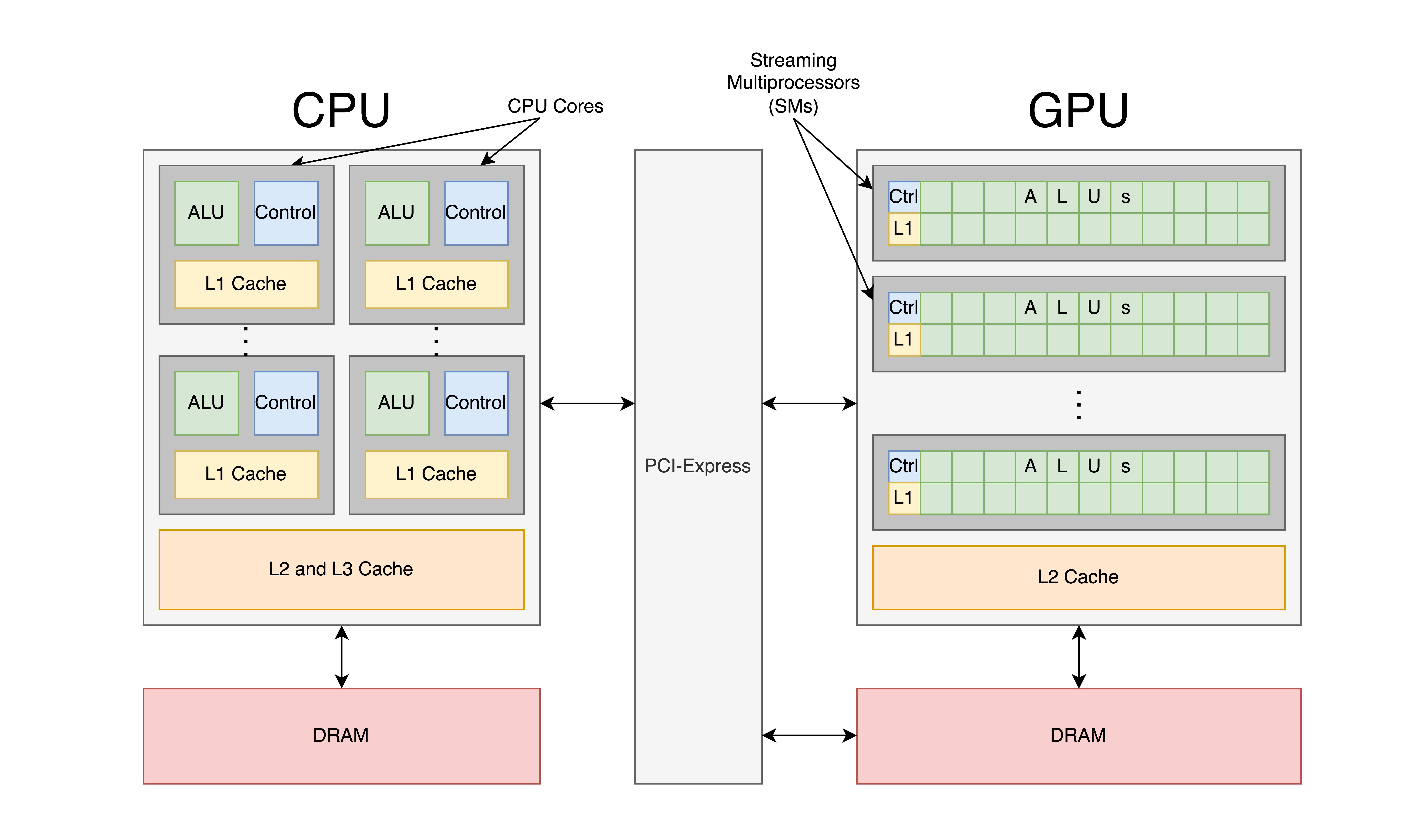}
    \caption{Comparison of the main differences in hardware architecture between CPUs (left) and GPUs (right).}
    \label{fig:cpu-vs-gpu}
\end{figure}

\subsection{Hardware Architecture}
GPUs are designed to implement the SIMD parallel computations required in graphics rendering (see Fig. \ref{fig:cpu-vs-gpu}) so much more of the hardware is focused on data processing as compared to central processing units (CPU) which have a more general processing architecture.(e.g. more emphasise on data caching and flow control). 

The Nvidia GPU architecture uses many streaming-multiprocessors (SM) which are comprised of many streaming processors or \textit{CUDA cores} and adopts an execution model called Single Instruction, Multiple Thread (SIMT), a variation of the SIMD architecture where the difference is that for SIMD a single instruction acts upon all data in the exact same way, whereas, in SIMT, selected threads can be activated or deactivated. This provides an effective mechanism for instructions and data to only be processed by active threads thus accommodating branching within the threads.

The properties and capabilities of Nvidia GPUs are identified by a version number, called the compute capability of the device, which can be accessed during run-time to determine and use features specific to a device. The first number of the compute capability determines the core architecture of the device and the minor version numbers correspond to incremental improvements to the core architecture. Table \ref{tab:compute_capability} summarises some selected features from compute capability 1.x to 9.0. A full technical overview can be found in \cite{ref:2022:nvidia_a}.

\subsection{Thread Hierarchy}
Many thousands of threads can execute within a kernel and are structured into a hierarchy to manage them. A \textit{grid} consists of all threads spawned by a kernel and are organised into \textit{thread blocks}. Thread blocks are processed by SMs and, depending on the number of threads, registers, and shared memory required, multiple thread blocks may execute on a SM. If sufficient resources are not available, thread blocks are queued until sufficient resources become available. The number of thread blocks per SM has varied over time, started 8 and currently has a maximum of 32 (see Table \ref{tab:compute_capability}). 

\begin{figure}[!t]
    \centering
    \includegraphics[width=\textwidth]{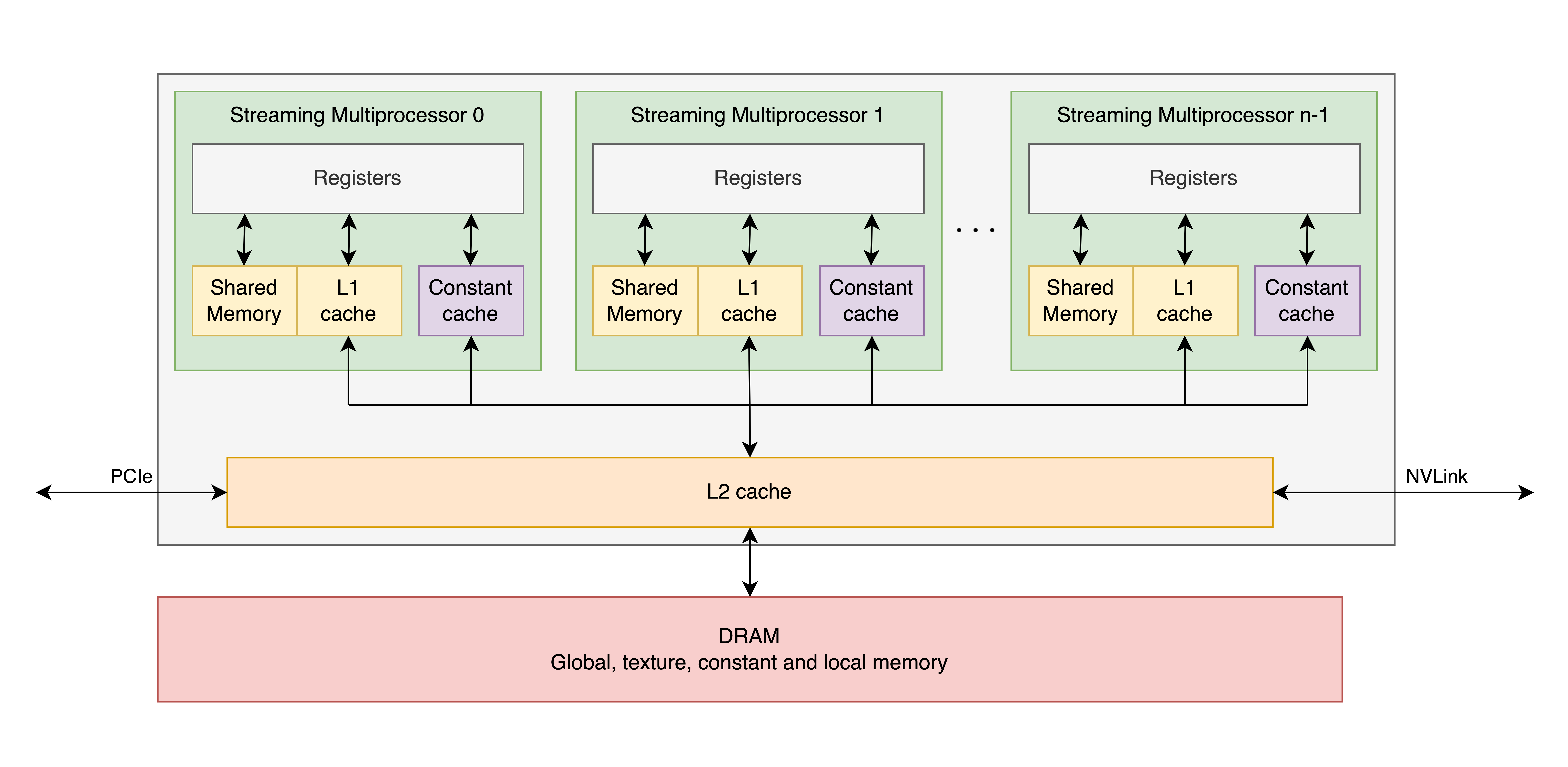}
    \caption{Memory hierarchy in Nvidia GPUs.}
    \label{fig:gpu_mem_levels}
\end{figure}

Communication between thread blocks can be achieved through off-chip DRAM referred to as \textit{global memory} (see Fig. \ref{fig:gpu_mem_levels}). No hardware synchronisation between thread blocks is possible and software synchronisation can lead to a deadlock as there is no explicit order in which thread blocks will be executed. However, devices with compute capability 9.0 have access to the concept of thread group clusters which enable multiple thread blocks running concurrently across multiple SMs to synchronise and exchange data \cite{ref:2022:nvidia_a}. Cooperation of threads within a thread block is achieved through intra-block synchronisation and the fast on-chip \textit{shared memory}. 

The threads of a thread block are divided into \textit{warps} of 32 threads and are processed one common instruction at a time in SIMT lock-step by 32 CUDA cores. When a divergent branch occurs within a warp, each branch taken must be executed sequentially using masking to enable and disable the various threads as required. Therefore, the threads of a warp should execute the same instruction to maximise throughput, whereas different warps execute independently regardless of flow paths. Each thread has access to its own \textit{registers} and \textit{local memory} to load and write to arbitrary memory addresses independently.

\subsection{Memory Hierarchy}
Threads in a CUDA program have access to multiple types of memory: 
\begin{itemize}
\item \textbf{Global}, \textbf{Local}, and \textbf{Texture} memories which are located in off-chip DRAM and have the greatest access latency; 
\item \textbf{Shared} memory which is located on-chip with much higher bandwidth and lower latency; 
\item \textbf{Constant} cache which is cached global memory; 
\item \textbf{Registers} which are extremely fast and local to the thread; and
\item \textbf{Cache} - L1 and L2 on-chip memory caches.
\end{itemize}
Only global, constant and texture memory are persistent across kernel launches within the same application. Fig. \ref{fig:gpu_mem_levels} depicts the different types of memory and how they are linked together. A more complete description of GPU memory architectures is given in the following sub-sections.

\subsubsection{Global memory}
All threads have access to global memory which is the most abundant memory on the GPU. When a warp executes an instruction that accesses global memory, the memory accesses of the warps threads are coalesced into one or more memory transactions. In general, as the number of transactions increases, the lower the throughput. To achieve the maximum performance, the threads of each warp should read blocks of contiguous global memory to produce the least number of memory transactions.

\subsubsection{Local memory}
Local memory is used to hold automatic variables when there is insufficient space to hold them in registers or L1 cache and is located in off-chip DRAM, making accesses just as expensive as global memory, however with a thread local scope.

\subsubsection{Texture memory}
Texture memory is cached, read-only memory which costs one read from the texture cache in L1 memory, or a device memory read during a cache miss. Texture memory is optimised for 2D spatial locality, therefore threads that read spatially close texture addresses achieve the best performance.

\subsubsection{Shared memory}
Shared memory is much faster than local and global memory, however it is only visible within a thread block, does not persist between kernel executions, and since compute capability 2.0 is limited to 48KB by default. Shared memory is physically in the same memory space as the L1 cache. Larger amounts of shared memory is available on some devices and must be manually requested, reducing the size of the L1 cache. Distributed shared memory is a new compute capability 9.0 feature allowing the threads of a thread block cluster access shared memory of another thread block. 

Shared memory is separated into 32 equally sized memory modules (banks) of 4 byte words that can be accessed simultaneously in order to obtain high memory bandwidth for concurrent accesses. This means that un-coalesced memory accesses in shared memory have no penalty. However, the accesses are serialised if multiple addresses in a memory request map to the same memory bank, resulting in a bank conflict, with the exception of a multi-cast which occurs when any number of threads in a warp access the same memory location.

\subsubsection{Constant memory}
Constant memory is read-only, limited to 64KB and is cached with 8KB per SM. A read from constant memory requires one read from the constant cache, or in the case of a cache miss, requires a read from device memory. Accesses to various addresses by threads inside a warp are serialised, therefore the cost increases linearly with the total number of distinct addresses read by warp threads. The constant cache works best when threads in the same warp only access a small number of distinct locations or if all threads of a warp access the same location.

\subsubsection{Registers}
Registers are directly accessed by the CUDA cores so are the fastest and are organised into 32 banks, matching the 32 threads of a warp. In general, reading a register requires no additional clock cycles for each instruction, although delays can happen due to register memory bank conflicts and read-after-write dependencies.  In order to prevent conflicts with register memory banks, the compiler and hardware thread scheduler will schedule instructions as efficiently as possible. 

Registers are thread local, however, using warp shuffle functions introduced with compute capability 3.0 the threads of a warp can exchange variables without the use of shared memory. The number of 32-bit registers per thread has increased from 124 in compute capability 1.x up to 255 from compute capability 3.2 onward.

\subsubsection{L1 Cache}
L1 cache is the on-chip storage location of cached global memory for each SM, providing fast access to recently read global memory data. The L1 cache is also used to manage register data that has overflowed an SMs register file.

\subsubsection{L2 Cache}
L2 cache is on-chip storage, however, it is shared by all SMs and is used to cache data transferred between the SMs and device memory. It is also used to mediate data transferred between host memory over PCI-E and multiple GPUs over NVlink.

\subsection{Random Number Generation} 
Prior to compute capability 2.0, random number generation was very difficult on the GPU and typically involved implementing a parallel pseudo-random number generator based on a linear congruential generator or Mersenne twister. However, since compute capability 2.0, Nvidia have developed the cuRAND library for pseudo-random and quasi-random number generation in CUDA kernels including uniform, normal, log normal, and Poisson distributions \cite{ref:2022:nvidia_b}.

% -----------------------------
% ---------- SECTION ----------
% -----------------------------
\section{Parallel models for GPU based Evolutionary Algorithms}
 The parallel models for GPU computing are extensions of the models developed for parallel EAs for CPU computing. Attempts to map EAs to GPUs have been made for over a decade and can be categorised into four categories that have been generalised from \cite{ref:2015:tan}: 
 \begin{itemize}
     \item \textbf{Naive parallel model} where the fitness evaluation is offloaded to the GPU; 
     \item \textbf{Multi-phase parallel model} where more obviously parallel aspects of the EA are offloaded to the GPU; 
     \item \textbf{All-GPU parallel model} where the entire EA is executed on the GPU; and 
     \item \textbf{Multi-population parallel model} in which multiple sub-populations of an EA are executed in parallel on the GPU. 
 \end{itemize}

The first three models are closely related to the master slave parallel model as the host executes compute kernels on the device, therefore offloading differing amounts of work to the GPU. The multi-population parallel model is also known as a GPU version of the coarse grained island model. The fine grained cellular model can be used in any of the four categories. Not to be confused with the parallel EA models, GPU implementations can use coarse grained parallelism where a thread is allocated to each individual for a \textit{task parallel} approach, or fine grained parallelism where multiple threads are allocated to each individual for a \textit{data parallel} approach.

\subsection{Naive parallel model}
The most computationally demanding aspect of an EA is the fitness evaluation, as all individuals in the population must be evaluated on a fitness function. As each fitness evaluation is independent, this is a perfect opportunity for parallelism and is the first bottleneck to be optimised in any EA. The type of objective function being optimised will determine what kind of parallelism is possible. This is a naive approach to GPU based parallelism of EAs. The earliest works in EA acceleration using GPUs offload the fitness evaluation to the GPU using a coarse grained approach, where the remainder of the algorithm was executed on the CPU \cite{ref:2009:maitre}. A benefit of the naive parallel model is any existing EA can be easily modified to perform the fitness evaluation on the GPU, while the remainder of the code can be left unchanged. A limitation is the PCI-Express bottleneck incurred when transferring the population and fitness values between the CPU and GPU every generation, especially when the EA is executed for thousands of iterations.

\subsection{Multi-phase parallel model}
To combat the limitation of the PCI-Express bottleneck of the naive parallel model, more of the EA can be moved to the GPU, therefore performing more work before moving the data back to the host. Earlier works struggled to move the entire DE algorithm to the GPU due to the limited random number generation of the GPU \cite{ref:2010:de}, however since compute capability 2.0 the cuRAND library \cite{ref:2022:nvidia_b} has been available allowing the generation of random numbers on the GPU.

\subsection{All-GPU parallel model}
To effectively remove the PCI-Express bottleneck and take advantage of the parallel computations provided by the GPU, the entire EA can be executed on the GPU, where only the final population or solution is transferred back to the host at the end of the algorithm. In this case, any remaining serial portions of code are moved to the GPU as the cost of PCI-Express transfer outweighs the cost of serial code execution. As phases of an EA need to complete before the next phase begins, synchronisation between operators must occur and no hardware synchronisation is available between thread blocks. To address the issue of synchronisation, one approach is to implement the entire algorithm using a single thread block, allocating one thread to each individual in the population. This limits the population size to the number of threads supported by a thread block which is 1024 as of compute capability 2.0 and forces all operations to use a task parallel approach, largely neglecting the inherent parallelism available. A second approach is to implement each operator as a separate kernel to provide synchronisation points, allowing both task parallel and data parallel approaches for each respective kernel as required. This is the most common implementation of GPU based EA, where each genetic operator is implemented as a separate kernel.

\subsection{Multi-Population parallel model} 
To take advantage of shared memory and reduce the use of global memory, the coarse grained island model can be executed on the GPU, where the island model maps to and utilises the GPU resources by allocating one \cite{ref:2010:luong, ref:2010:pospichal} or many \cite{ref:2022:janssen} threads to each individual in the population and allocating each sub-population to a thread block. This allows synchronisation within the thread block and reuse of shared memory for many iterations. Synchronous and Asynchronous migration approaches exist: the synchronous approach will end the kernel when a migration is set to occur using a migration kernel; and the asynchronous approach will perform the migration in the same kernel. The asynchronous approach has the advantage of a single kernel, however, due to thread block scheduling,  unintended individuals may be migrated. Thread blocks are executed as GPU resources become available and any time the number of thread blocks exceeds the resources of the GPU, the excess thread blocks will be queued. This means asynchronous migration can lead to the earliest thread blocks to execute may migrate with islands that have not begun execution, or later thread blocks may migrate with islands that have already completed execution \cite{ref:2022:janssen}. The only time the kernel is stopped is either a stopping criteria has been met, or a migration is set to occur through global memory using a migration kernel.

% -----------------------------
% ---------- SECTION ----------
% -----------------------------
\section{Differential Evolution on GPUs}

DE is a relatively simple EA optimisation algorithm as each parameter vector within a solution can be modified by the mutation and crossover operators independently. This makes DE a good candidate for implementation using GPUs. To date, the approaches taken can be classified as using the multi-phase parallel model or All-GPU parallel model. Although numerous GPU accelerated DE algorithms exist, it is almost impossible to compare them effectively because:
\begin{itemize}
\item Only one of the current algorithms is completely open source \cite{ref:2020:mateus} however partial source code is available for \cite{ref:2010:de};
\item The differences between hardware and compute capability used to run the experiments; and perhaps most importantly, 
\item There is no standard set of benchmark problems for evaluating the performance of each implementation. 
\end{itemize}

All examined works compare their GPU based implementation against a single-threaded CPU implementation and discuss the provided speed-ups, but as is also the case with GPU hardware, the actual CPUs used range drastically. Another consideration is that commonly the CPU implementations are single-threaded, providing a one-sided comparison when comparing against a massively parallel architecture such as a GPU. Including multi-threaded CPU implementations into the comparisons could provide further insight into the benefit of GPU based DE, or perhaps cause some of the reported speed-ups to be less appealing.

\subsection{Kernel Design} 
To implement DE, most GPU based approaches separate the algorithm into the following steps: 

\begin{enumerate}
    \item Population Initialisation
    \item Function Evaluation
    \item Trial Vector Selection
    \item Trial Vector Generation
    \item Replacement
\end{enumerate}

\begin{figure}[!t]
    \centering
    \includegraphics[width=\textwidth]{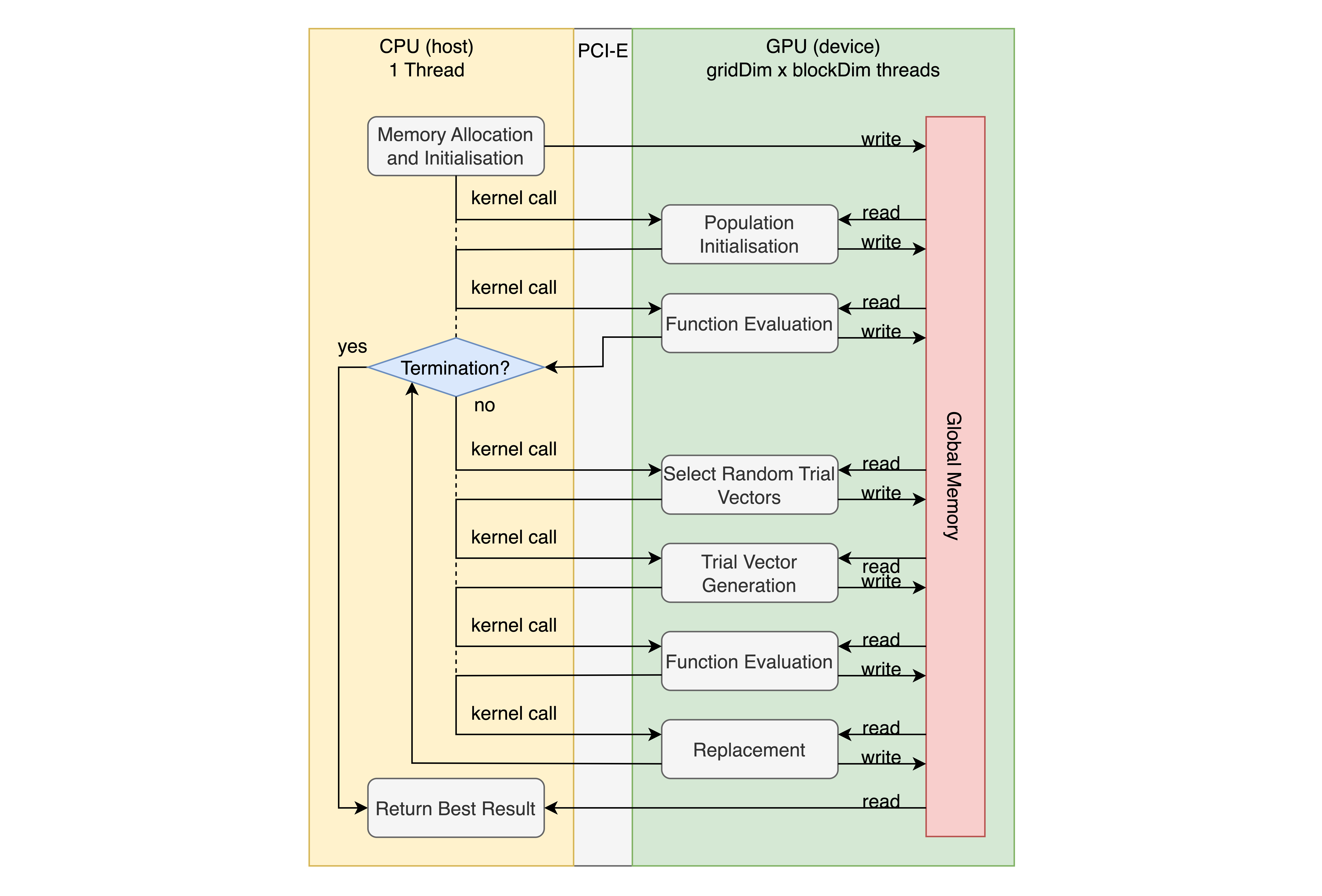}
    \caption{DE using All-GPU parallel model where each DE operation is implemented as a separate kernel.}
    \label{fig:2011_de}
\end{figure}

These five steps have different possible parallel grain for their implementation. Population Initialisation, Function Evaluation, and Trial Vector Generation can be implemented using fine grain parallelism - allocating multiple threads to each individual to accelerate the operations. Whereas, Trial Vector Selection and Replacement are coarse grain task parallel - allocating a single thread to each individual. Replacement may be implemented using fine grain parallelism if the parameter data needs to be copied, rather than swapping the pointers to the data. 

The first GPU based DE algorithm used the \textbf{Multi-phase parallel model} due to the limitations of compute capability 1.x, where \cite{ref:2010:de} implemented the Trial Vector Selection on the CPU, where the remainder of the algorithm was implemented on the GPU. However, most GPU based DE algorithms now use the \textbf{All-GPU parallel model}, where the whole algorithm executes on the GPU, and each of the previously defined operations are implemented as CUDA kernels as shown in Fig. \ref{fig:2011_de}. This is the general structure used in \cite{ref:2010:de} \cite{ref:2011:kromer} \cite{ref:2011:kromer_b}. Additional kernels are added for operations required in more advanced variations of DE such as GPU based SaDE (cuSaDE)\cite{ref:2015:wong} and GPU based jDE (cujDE)\cite{ref:2020:mateus}, which add an additional kernel for learning, or parameter updating, while following the same overall structure. 

\begin{figure}[!t]
    \centering
    \includegraphics[width=\textwidth]{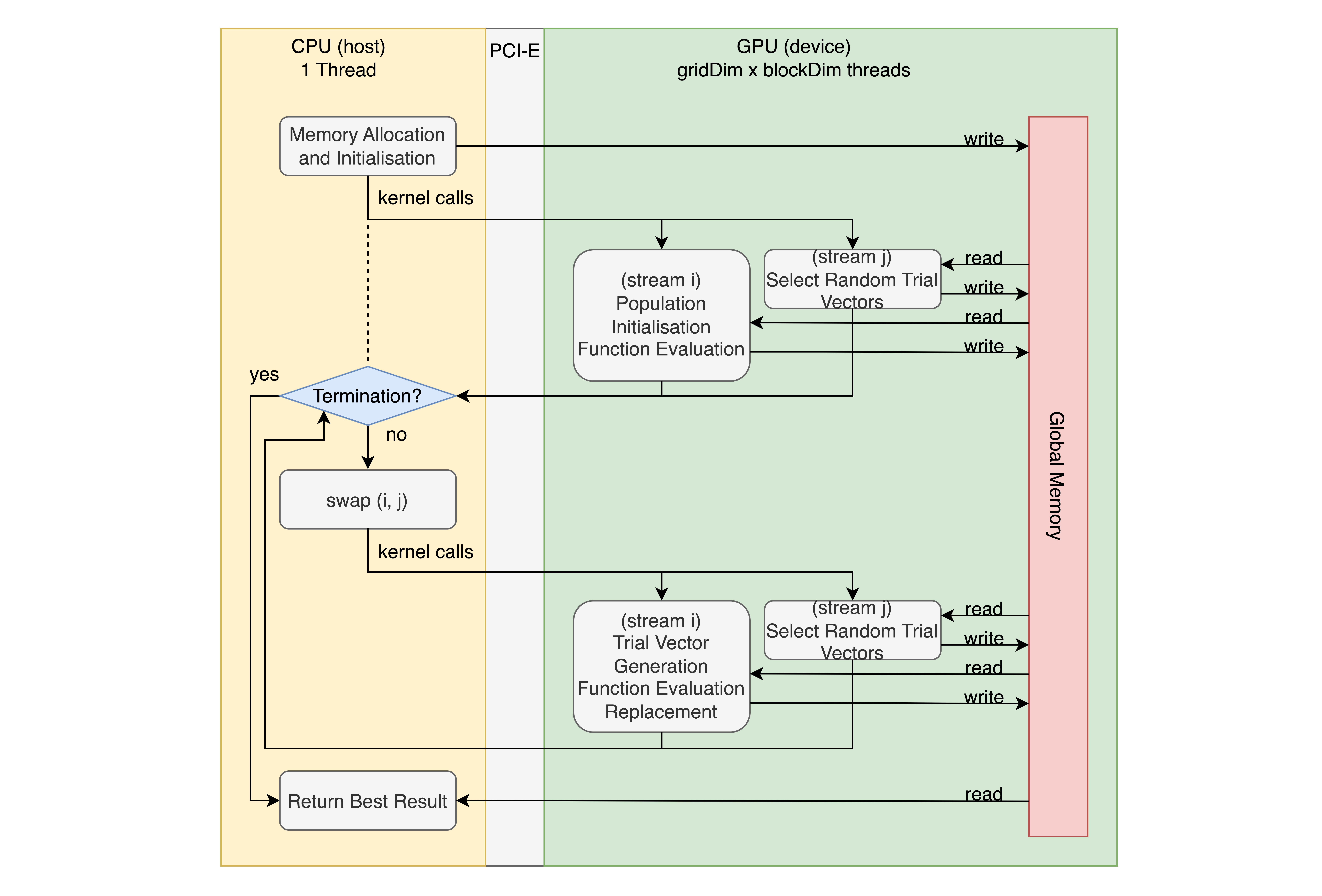}
    \caption{Improved DE using All-GPU parallel model that combined several logically-related kernels to increase shared memory use, reduce global memory access, and utilised CUDA streams to enable concurrent kernel execution. }
    \label{fig:2012_de}
\end{figure}

In 2012, an improved All-GPU based DE (iDE) was presented \cite{ref:2012:qin}, addressing the issues of low-throughput memory access and less efficient device utilisation when using separate kernels for each DE operation. As DE is run for many iterations, given the previously defined structure, each kernel is executed hundreds, or thousands of times, where each of the kernels load data to and from global memory. 

The All-GPU iDE optimised memory and device utilisation by combining several logically-related kernels to reduce global memory access and maximise the use of shared memory, determining kernel execution parameters automatically to maximise device occupancy of the GPU, and to maximise device utilisation CUDA streams were used to enable concurrent kernel execution as shown in Fig. \ref{fig:2012_de}. However, this approach still requires global memory reads every iteration.

Currently, no existing GPU based DE algorithm use the \textbf{Multi-population parallel model}.

\subsection{Differential Evolution Operations}
This section discusses the different implementation approaches for implementing GPU based DE operations.

\subsubsection{Population Initialisation}
Since the earliest works population initialisation has been performed on the GPU. Originally GPU based DE used a Mersenne twister provided by the Nvidia CUDA SDK for population initialisation \cite{ref:2010:de}, this was quickly replaced by the cuRAND library \cite{ref:2011:kromer} \cite{ref:2012:qin} \cite{ref:2015:wong}. These approaches allocate a thread for each parameter of every individual and generate the random population in one step in parallel. The most recent work utilises the thrust library to generate the initial population \cite{ref:2020:mateus}. 

\subsubsection{Function Evaluation} 
Function evaluation is highly dependent on the problem being optimised, an overview of functions and problems that GPU-based DE has been used to optimise is included in the benchmark discussion. 

\subsubsection{Trial Vector Selection}
A requirement when implementing DE in parallel is the generation of the mutually exclusive random integers used for the random trial vectors required for generating the mutant vectors. The traditional method is using a trial-and-error loop. 

Originally, random number generation on the GPU was difficult as the cuRAND library was not yet available. Due to this limitation, the earliest work used the \textit{rand} function of the C standard library on the host for generating the mutually exclusive random integers required for the \textit{rand/1/bin} strategy, transferring them to the device at each iteration \cite{ref:2010:de}. 

Once the cuRAND library became available, generation of the mutually exclusive integers could be performed by a kernel, allocating one thread per population member for coarse grain parallelism \cite{ref:2010:de} \cite{ref:2015:wong}. 

To better utilise the GPU hardware, \cite{ref:2012:qin} performed the selection of random trial vectors concurrently with another kernel using CUDA streams, generating the mutually exclusive random numbers for the next generation while concurrently processing the current generation. 

Instead of using the trial-and-error loop, \cite{ref:2020:mateus} performed generation of the mutually exclusive integers required for the \textit{rand/1/bin} strategy by randomly generating three displacement values with the first in the range $[1, N_P]$ and the remaining two in the range $[1, \frac{N_P}{3}]$, where to obtain the three random values, a partial sum was performed. This ensured that only three random numbers needed to be generated, whereas using the trial-and-error loop many random numbers may be required when the same integer is generated multiple times. 

\subsubsection{Trial Vector Generation}
To generate the trial vectors, \cite{ref:2010:de} launched 32 blocks of 64 threads where each thread block processed $\frac{N_P}{32}$ individuals, and each thread processed $\frac{D}{64}$ parameters of each individual. The threads load their respective trial vector indices from global memory into local variables, and generate the trial vector in global memory using a parallel loop. 

Ref. \cite{ref:2011:kromer}\cite{ref:2011:kromer_b} used a similar approach however launching 1024 threads per block, allowing the entire population to be processed in one step whereas \cite{ref:2015:wong} loaded the trial vector indices, and other parameters required for SaDE from global memory into shared memory for fast access. An integer multiple of the problem dimension threads were allocated per block, and each thread handled one parameter of one individual. This ensured that any trial vector is entirely generated by the same thread block. Similarly,  \cite{ref:2020:mateus} loaded the trial vector indices, and other parameters required for jDE from global memory into shared memory, however allocating a block of the next power of two greater than the problem dimension threads per individual. 

\subsubsection{Replacement}
Replacement requires a comparison between the parent and offspring individuals, where the better individual will continue in the next generation. In \cite{ref:2010:de} the replacement operation is combined with the function evaluation, where 32 blocks of 64 threads are launched to perform the operations using a fine grained approach. 

In \cite{ref:2011:kromer}\cite{ref:2011:kromer_b} a kernel was implemented for the replacement operator, launching 1024 threads per block to process and merge the parent and offspring population in one step while \cite{ref:2015:wong} chose a task parallel approach, allocating one thread per individual to compare each individual with its respective trial vector, writing the fitter one and its corresponding fitness value into global memory. This kernel also updated the success and failure counts, as well as the successful $C_R$ values required for the learning operation in SaDE using atomic operations. 

Ref. \cite{ref:2020:mateus} also used a single thread per individual task parallel approach, writing the fitter individual into global memory. An obvious issue with \cite{ref:2015:wong}\cite{ref:2020:mateus} is the task-parallel approach of allocating a single thread to perform the replacement operation, as a better offspring overwrites the parameter vector of the parent this operation can be easily implemented in a fine grained approach as performed in \cite{ref:2010:de}\cite{ref:2011:kromer}\cite{ref:2011:kromer_b}.

% -----------------------------
% ---------- SECTION ----------
% -----------------------------
\section{GPU Based DE Benchmarks}
This section first reviews the diverse range of results and benchmark problems GPU based DE has been used for, followed by a discussion in numerical optimisation benchmarks, and finally a new GPU based numerical optimisation benchmark is presented.

\subsection{Current Benchmarking Environment}
Currently the benchmarks used to evaluate GPU based DEs makes comparisons of algorithms difficult. Some of the current evaluation techniques are:
\begin{itemize}
\item Ref. \cite{ref:2010:de} evaluated on six bound constrained 100-dimensional benchmark functions with various lower and upper bounds \cite{ref:2004:chang} using 100 individuals over 10,000 iterations, and 1000 individuals over 100,000 iterations. Each experiment was run 20 times and compared against an equivalent single-threaded C implementation. The maximum speedup achieved with 100 individuals was 19 times, and with 1000 individuals achieving a maximum speedup of 34 times over the single-threaded C implementation.

\item Ref. \cite{ref:2011:kromer} evaluated on two combinatorial optimisation problems: Task scheduling; and Linear ordering, as well as one of the numerical optimisation problems from \cite{ref:2010:de}. It achieved between 2.2 times to 12.5 times faster than a single threaded performance focused C implementation for the task scheduling problem for 512 jobs and 16 machines. The 50-dimensional linear ordering problem (LOP) compared the speed of evaluation between CPU and GPU using population sizes ranging from 16 to 32768, increasing in powers of two. They found that the GPU implementation began to outperform the CPU at 256 individuals onward, finding a speedup of over two times. A comparison with \cite{ref:2010:de} for one of the same numerical optimisation problems was made, however, 128 dimensions were used rather than 100 and this algorithm was found to be competitive with the previous CUDA-C implementation. 

\item Ref. \cite{ref:2012:qin} compared their All-GPU iDE against three DE algorithms implementing the \textit{rand/1/bin} strategy: a CPU based sequential DE and two GPU based DE implementations \cite{ref:2010:de} \cite{ref:2011:gonzalez} on four numerical optimisation problems from the CEC2005 special session on real-parameter optimization \cite{ref:2005:suganthan}: shifted sphere; shifted Rosenbrock's; shifted Griewank's; and shifted Rastrigin's functions. Experiments with population sizes 50, 100, 500, and 1000, with problem dimensions 10, 50, 100 and a termination criteria of $10^4\times D$ function evaluations were performed. Each experiment was run 25 times with different random seeds, however with the same initial starting seed between algorithms, and set the parameters $C_R=0.3$ and $F=0.5$ for all implementations. For all test problems their implementation consistently demonstrated superior computational time efficiency for all population sizes and problem dimensions.

\item Ref. \cite{ref:2015:wong} test their GPU based SaDE algorithm on four numerical optimisation problems selected from the CEC 2013 special session and competition on real-parameter optimisation \cite{ref:2013:cec2013}: shifted sphere; rotated Rosenbrock's; rotated Griewank's; and rotated Rastrigin's functions. The learning period $LP$ was set to 50 and experiments with population sizes 30, 50, 100, 300, 500, and problem dimensions 10, 50, 100 were performed. Each experiment was run 15 times with different seeds but used the same initial seed between hyperparameters and used the CEC 2013 stopping criteria of $D\times10^4$ function evaluations. In summary, this implementation had similar optimisation accuracy to the CPU counterpart and found that high dimension and more complicated objective functions had larger speed-ups when compared to the CPU implementation. 

\item Ref. \cite{ref:2020:mateus} evaluated their GPU based jDE algorithm cujDE on the same four numerical optimisation problems as \cite{ref:2015:wong} from the CEC 2013 benchmark: shifted sphere; rotated Rosenbrock's; rotated Griewank's; and rotated Rastrigin's functions, at 50 and 100 dimensions with population sizes 50, 100, 300, 500. Each experiment was run 15 times with different random seeds and terminated the search after $D\times10^4$ function evaluations. This work compared their algorithm against a CPU based sequential algorithm and compared the optimisation accuracy with the results obtained in \cite{ref:2015:wong}. Compared to the sequential CPU implementation a speedup of 48.4 times was achieved with similar optimisation accuracy between the implementations. The algorithm performed competitively against \cite{ref:2015:wong}, equivalently solving shifted sphere, performing better in rotated Griewank's and rotated Rastrigin's functions, and performing worse in rotated Rosenbrock's function. cujDE has since been applied to protein structure prediction in \cite{ref:2019:boiani}\cite{ref:2020:boiani}\cite{ref:2021:parpinelli}.
\end{itemize}

\subsection{Numerical Optimisation Benchmarks}
A lot of work has been conducted in GPU based DE, where the earlier works sought to improve upon the wall-clock time performance of the regular DE algorithm, optimising the implementation to maximise SM performance \cite{ref:2012:qin}. Whereas the later works instead chose the simpler kernel for each operation approach, however selecting variants of DE that improve convergence properties \cite{ref:2015:wong, ref:2020:mateus}. Determining the relative performance of an algorithm can be difficult when there are no standardised set of experiments freely available to run. 

As seen in the previous section, the majority of the GPU based DE algorithms were evaluated on differing problems of varying dimension with various population sizes, comparing with equivalent CPU based algorithms where the focus was largely on computational speedup where convergence properties equivalent with the CPU counterpart were desired. A GPU based benchmark needs to be developed to allow these algorithms to be compared and determine which are competitive with respect to both computational speedup and convergence properties. 

Currently, for numerical optimisation algorithms, the benchmark suites from the Congress on Evolutionary Computation (CEC) competitions are commonly used to evaluate the quality of a proposed algorithm \cite{ref:2005:cec2005, ref:2013:cec2013, ref:2014:cec2014, ref:2017:cec2017, ref:2020:cec2020, ref:2021:cec2021, ref:2022:cec2022}. These benchmark suites are comprised of unimodal, basic, hybrid, and composition bound constrained numerical optimisation functions in various dimensions with set computational budgets. 

The benchmark functions often have transformations applied, such as rotation and shift. Shift is applied to randomise the global optimum's location in the search space to prevent centre-biased algorithms from finding solutions on the origin, and rotation is applied to remove linkages between variables. The shift and rotation values are provided to ensure each benchmark evaluation is consistent. 

The computational budget applied to the CEC benchmarks is limiting the number of function evaluations an algorithm can perform before termination. This is a good metric when used to ensure that algorithms are not run for excessively long amounts of time to produce good solutions. The maximum number of function evaluations have changed between CEC competitions, where the CEC'13, CEC'14, and CEC'17 competitions allowed $D\times 10^4$ function evaluations, and the more recent competitions reduced the search dimensions to a maximum of $D=20$ with much larger computational budgets: CEC'20 allowed 50,000, 1,000,000, 3,000,000, and 10,000,000 function evaluations for dimensions D=5, D=10, D=15, and D=20, respectively, and CEC'21 and CEC'22 allowed 200,000, and 1,000,000 function evaluations for dimensions D=10, and D=20, respectively. 

The latest competitions have also reduced the number of functions contained in the benchmark suites, with a focus on hybrid and composition functions, where the functions have been selected from the CEC'14 and CEC'17 benchmarks. In all competitions early stopping is applied if an error value of $10^{-8}$ is reached before exhausting the computational budget. 

As the CEC set of benchmark functions change between the competitions, it has been noted that competing algorithms are only compared with other algorithms of the same year and not with algorithms from previous years \cite{ref:2017:molina, ref:2019:skvorc}. It was shown that the algorithms that perform the best in one competition might not necessarily perform better than the algorithms presented in previous years. This presents an important consideration when comparing algorithms, without running the exact same set of benchmark functions to evaluate the performance of a set of algorithms, the conclusions drawn can be misleading. This can occur when using what appear to be the same functions, as the CEC competitions perform shift and rotation on the functions, the shift and rotate values and methods have changed between competitions.

\begin{table*}[!t]
\caption{Proposed GPU Based Numerical Optimisation Benchmark Comprised of 1 Uni-modal, 3 Basic, 3 Hybrid, and 3 Composition Functions. }
\label{tab:benchmark_proposal}
\centering
\resizebox{\textwidth}{!}
{
\begin{tabular}{cccclc}
\hline
\textbf{Function Type} & \textbf{No.} & \textbf{Functions} & \textbf{Hybrid and Composition Basic Functions} & \textbf{Parameters} & \textbf{Competition} \\ \hline
Uni-modal & 1 & Zakharov Function &  &  & CEC'17 F03 \\ \hline
\multirow{3}{*}{Basic} & 2 & Rosenbrock's Function &  &  & CEC'17 F04 \\ \cline{2-6} 
 & 3 & Rastrigin's Function &  &  & CEC'17 F05 \\ \cline{2-6} 
 & 4 & Schwefel's Function &  &  & CEC'17 F10 \\ \hline
\multirow{12}{*}{Hybrid} & \multirow{3}{*}{5} & \multirow{3}{*}{Hybrid Function 1} & Bent Cigar Function & \multirow{3}{*}{$p=[0.3, 0.3, 0.4]$} & \multirow{3}{*}{CEC'14 F18} \\ \cline{4-4}
 &  &  & HGBat Function &  &  \\ \cline{4-4}
 &  &  & Rastrigin's Function &  &  \\ \cline{2-6} 
 & \multirow{4}{*}{6} & \multirow{4}{*}{Hybrid Function 2} & Expanded Schaffer F6   Function & \multirow{4}{*}{$p=[0.2,0.2,0.3,0.3]$} & \multirow{4}{*}{CEC'17 F16} \\ \cline{4-4}
 &  &  & HGBat Function &  &  \\ \cline{4-4}
 &  &  & Rosenbrock's Function &  &  \\ \cline{4-4}
 &  &  & Modified Schwefel's Function &  &  \\ \cline{2-6} 
 & \multirow{5}{*}{7} & \multirow{5}{*}{Hybrid Function 3} & Katsuura Function & \multirow{5}{*}{$p=[0.3,0.2,0.2,0.1,0.2]$} & \multirow{5}{*}{CEC'14 F22} \\ \cline{4-4}
 &  &  & HappyCat Function &  &  \\ \cline{4-4}
 &  &  & Expanded Griewank's plus Rosenbrock's Function &  &  \\ \cline{4-4}
 &  &  & Modified Schwefel's Function &  &  \\ \cline{4-4}
 &  &  & Ackley's Function &  &  \\ \hline
\multirow{12}{*}{Composition} & \multirow{3}{*}{8} & \multirow{3}{*}{Composition Function 1} & Rastrigin's Function & \multirow{3}{*}{\begin{tabular}[c]{@{}l@{}}$\sigma=[10,20,30]$\\ $\lambda=[1,10,1]$\\ ${bias}=[0,100,200]$\end{tabular}} & \multirow{3}{*}{CEC'17 F22} \\ \cline{4-4}
 &  &  & Griewank's Function &  &  \\ \cline{4-4}
 &  &  & Modified Schwefel's Function &  &  \\ \cline{2-6} 
 & \multirow{4}{*}{9} & \multirow{4}{*}{Composition Function 2} & Ackley's Function & \multirow{4}{*}{\begin{tabular}[c]{@{}l@{}}$\sigma=[10,20,30,40]$\\ $\lambda=[10,1e-6,10,1]$\\ ${bias}=[0,100,200,300]$\end{tabular}} & \multirow{4}{*}{CEC'17 F24} \\ \cline{4-4}
 &  &  & High Conditioned Elliptic Function &  &  \\ \cline{4-4}
 &  &  & Griewank's Function &  &  \\ \cline{4-4}
 &  &  & Rastrigin's Function &  &  \\ \cline{2-6} 
 & \multirow{5}{*}{10} & \multirow{5}{*}{Composition Function 3} & Expanded Schaffer F6   Function & \multirow{5}{*}{\begin{tabular}[c]{@{}l@{}}$\sigma=[10,20,20,30,40]$\\ $\lambda=[0.005,1,10,1,10]$\\ ${bias}=[0,100,200,300,400]$\end{tabular}} & \multirow{5}{*}{CEC'17 F26} \\ \cline{4-4}
 &  &  & Modified Schwefel's Function &  &  \\ \cline{4-4}
 &  &  & Griewank's Function &  &  \\ \cline{4-4}
 &  &  & Rosenbrock's Function &  &  \\ \cline{4-4}
 &  &  & Rastrigin's Function &  &  \\ \hline
\end{tabular}
}
\end{table*}

\subsection{Benchmark Proposal}
It is proposed that a set of functions be selected and that implementations for GPU based algorithms are available to generate a competitive benchmark for comparison between GPU based algorithms. Similar to the CEC'20, CEC'21, and CEC'22 benchmark suites, a selection of functions have been chosen from CEC'14 and CEC'17 for the GPU based benchmark with emphasis on hybrid and composition functions. As these competitions contained 50 and 100 dimensional problems, the input files provided can be used for shift, rotate, and shuffle for the GPU based benchmark. Table \ref{tab:benchmark_proposal} contains the proposed benchmark consisting of 1 uni-modal, 3 basic, 3 hybrid, and 3 composition functions, where the 3 hybrid and 3 composition functions are composed of 3, 4, and 5 basic functions, respectively. The table includes the originating competition for each function from which the respective shift, rotate, or shuffle data is obtained from.

\subsubsection{Function descriptions} 
All of the benchmark functions are minimisation functions, defined as: 

\[ \min f(x), x = [x_1, x_2, \ldots, x_D] \] 

where $D$ is the dimension of the problem. The search range is fixed to $[-100, 100]$ for all dimensions and each function scales the search range down to the correct domain of the respective objective function. The global minimum for all objective functions is defined at $x=[0, 0, \ldots, 0]$, however to avoid centre biased algorithms from exploiting the solution on the origin, a shift vector $o=[o_1, o_2, \ldots, o_D]$ is provided for each function to shift the global optima to a randomised position between $[-80, 80]$ for each problem dimension, giving the shifted parameter vector $y = x - o$. To reduce linkages between the variables of $y$, a rotation matrix $M$ is then applied to the scaled and shifted parameter vector where a different rotation matrix is provided for each function to generate the final parameter vector $z = My$ to be evaluated on the benchmark function.

\subsubsection{Hybrid functions} 
The hybrid functions first shift and rotate the parameter vector, followed by shuffling the resultant vector using a provided random permutation, and finally splitting the parameter vector into multiple sub components, where each sub component is evaluated on a different basic function. The percentage of each sub component is defined for each hybrid function. The hybrid functions are defined as: 

\[ f(x) = g_1(M_1z_1) + g_2(M_2z_2) + \ldots + g_n(M_nz_n) \]

where $f(x)$ is the hybrid function, $g_i(x)$ is the $i^{th}$ basic function, $n$ is the number of sub components, $z$ is a random permutation of the shifted and rotated parameter vector $x$ and $z_i$ is a sub-component of $z$.

\subsubsection{Composition functions} 
Composition functions are combinations of multiple basic functions, each with a separate weighting which merges properties of the basic functions, maintaining continuity around a global optimum. The local optimum with the smallest bias value becomes the global optimum. The composition functions are defined as: 

\[ f(x) = \sum_{i=1}^{n} \omega_i[\lambda_ig_i(x)+{bias}_i] \]

where $f(x)$ is the composition function, $g_i(x)$ is the $i^{th}$ basic function, $n$ is the number of basic functions, $\lambda_i$ is used to control the height of $g_i(x)$, ${bias}_i$ defines the global optimum and $\omega_i$ is the weight value of $g_i$. The weight value is calculated as follows: 

\[ w_i = \frac{1}{\sqrt{\sum_{j=1}^{D}(x_j-o_{ij})^2}} \exp{- \frac{\sum_{j=1}^{D}(x_j-o_{ij})^2}{2D\sigma_i^2}} \]

where $o$ is the shift vector for the $i^{th}$ basic function and $\sigma_i$ is used to control the coverage range of $g_i$. The weight must then be normalised: 

\[ \omega_i = \frac{w_i}{\sum_{i=1}^{n} w_i} \]

All basic functions are shifted and rotated, therefore each composition function requires $n$ shift vectors and $n$ rotation matrices.

\subsubsection{Stopping criteria}
A major limitation imposed by earlier CEC competitions was the computational budget, however the CEC22 benchmark significantly increased the budget to 200,000 for $D=10$ and 1,000,000 for $D=20$. 

For the proposed benchmark, a maximum number of function evaluations of 5,000,000 for $D=50$ and 10,000,000 for $D=100$ should be sufficient to allow for algorithms that use larger populations to effectively explore the search space. This is necessary as the primary motivation for utilising GPU computation is the massively parallel computation which can be better harnessed with larger population sizes. Each experiment should be run 30 times, and the minimum error value and wall-clock time should be recorded either once the maximum function evaluations is exhausted or an error value of $10^{-8}$ is reached. If early stopping is achieved, the number of function evaluations used should be recorded. 

In addition and, in contrast to previous GPU based algorithms that explored different population sizes for different problems, all parameters should be fixed for the benchmark, participants are not allowed to search for distinct parameters for each problem or dimension.

\subsubsection{Evaluation criteria}
The evaluation criteria is similar to that presented in the CEC'22 competition. The evaluation criteria rewards both speed and accuracy, acknowledging that rather than relying on the mean values obtained by the multiple trials of an algorithm, trials can be ranked from best to worst when they reach the minimum error value or the maximum number of function evaluations. For each algorithm, the respective trial ranks can be compared, summed, and a correction term is subtracted. This way the algorithms are more fairly compared, and outlying trials are less likely to skew results, this test can be interpreted as comparing the median between two populations of results. 

To compare algorithms, the CEC'22 competition ranked trials from multiple algorithms from best to worst when they terminate upon reaching a minimum error value or the maximum number of function evaluations. In contrast to this, the proposed evaluation uses a similar ranking system, replacing the function evaluations with wall-clock time emphasising the computational speedup provided by GPUs, without the limitation of function evaluations. For example, allowing large populations that are evaluated in parallel to perform competitively against algorithms which reduce the population size during run-time, with respect to the wall-clock run-time. Using the evaluation criteria, parameters such as population size can be explored and compared for the benchmark. 

To calculate the algorithm score for a given function, rank all of the trials for all competing algorithms where ${trial}_{i,j}$ is the $i^{th}$ trial from the $j^{th}$ algorithm, $i=1,2,\ldots,n$, $j=1,2,\ldots,m$, where $n$ is the number of trials, and $m$ is the number of competing algorithms. The best trial is assigned the highest rank of $nm$, and tied trials are assigned an average rank. The final algorithm score for the given function is the sum of its ranks minus the correction term $\frac{n(n+1)}{2}$. To produce the final score over all functions in the benchmark all of the function scores are summed.

% -----------------------------
% ---------- SECTION ----------
% -----------------------------
\section{Case Studies} 
Two case studies are presented, the first compares the computational speedup provided by general purpose GPU computation and the second demonstrates the evaluation criteria to compare an All-GPU parallel model and the open source cujDE algorithm. The experiments were conducted on a PC equipped with an Intel i9-10900F CPU @ 2.80GHz (5.20GHz boost) with 10 cores supporting 20 threads, and an NVIDIA GeForce RTX 3080 GPU @ 1.44GHz (1.71GHz boost) with 68 Amphere SMs with a total of 8704 CUDA cores, 10GB of GDDR6X global memory, and compute capability 8.6. The development environment was Ubuntu 22.04 operating system and CUDA version 12.1. 

The GPU based benchmark was implemented so that a block of threads are allocated for each individual in the population, where the number of threads per block is a power of two greater than the problem dimension. Two arrays of shared memory are required for each block to facilitate the temporary memory required to shift, scale, and rotate the individuals parameter vector loaded from global memory and then used to evaluate the objective value. The size of the arrays is the next power of two greater than the problem dimension to prevent bank conflicts in shared memory. The rotation matrices are transposed to allow coalesced memory access when rotating the parameter vector.

\subsection{Case study 1}
To demonstrate the proposed benchmark and emphasise the benefit of GPU computation, a comparison is made using a standard DE algorithm with the same parameters in various configurations: 

 \begin{itemize}
     \item \textbf{Standard DE} which is the regular single-threaded DE algorithm;
     \item \textbf{Master Slave DE} which offloads fitness evaluation to a pool of CPU threads; 
     \item \textbf{Naive Parallel DE} which offloads fitness evaluation to the GPU;
     \item \textbf{All-GPU DE} where the entire DE algorithm is executed on the GPU, using the multiple kernel approach outlined in Fig. \ref{fig:2011_de}.
     \item \textbf{All-GPU iDE} where the entire DE algorithm is executed on the GPU, using the improved DE (iDE) approach outlined in Fig. \ref{fig:2012_de}.
 \end{itemize}

All implementations check whether a solution has been found each iteration. The CPU based models (standard DE, Master Slave DE, and Naive Parallel DE) use a linear search of the fitness values to determine if a solution has been found. The GPU based models (All-GPU DE and All-GPU iDE) use a GPU based approach, where the first thread of the block responsible for an individual will check the fitness value, and if the fitness value is below $10^{-8}$ an atomic operation is used to set a Boolean flag to indicate a solution has been found and store the index of the respective individual in global memory. An additional CUDA stream was used to asynchronously copy the data to the host during each iteration to check whether a solution had been found. 

Both the All-GPU DE and All-GPU iDE allocate a block of threads to each individual, with the number of threads being a power of two greater than the problem dimension. The All-GPU iDE uses additional shared memory to store the trial vector for faster memory access. 

\begin{table*}[!t]
\caption{Average Time in Seconds over 30 Trials on each 100-dimensional Benchmark Function with the Five Different Implementations to Exhaust the 10,000,000 Function Evaluation Budget. The Speedup Value is the Speedup Obtained Comparing the Different Implementations Against the Standard DE on each Respective Function. }
\label{tab:benchmark_compare}
\centering
\begin{tabular}{lllllllllll}
\hline
 & \textbf{F01} & \textbf{F02} & \textbf{F03} & \textbf{F04} & \textbf{F05} & \textbf{F06} & \textbf{F07} & \textbf{F08} & \textbf{F09} & \textbf{F10} \\ \hline
\textbf{Standard DE   (CPU)} & 80.81 & 80.74 & 94.30 & 112.62 & 86.88 & 94.23 & 190.34 & 277.73 & 329.80 & 434.04 \\ \hline
Speedup & 1.00 & 1.00 & 1.00 & 1.00 & 1.00 & 1.00 & 1.00 & 1.00 & 1.00 & 1.00 \\ \hline
\textbf{Master Slave DE (CPU)} & 22.046 & 21.881 & 23.359 & 25.729 & 22.714 & 23.438 & 34.604 & 41.619 & 45.582 & 55.454 \\ \hline
Speedup & 3.67 & 3.69 & 4.04 & 4.38 & 3.82 & 4.02 & 5.50 & 6.67 & 7.24 & 7.83 \\ \hline
\textbf{Naïve Parallel DE (CPU/GPU)} & 14.35 & 14.01 & 14.09 & 14.53 & 14.40 & 14.52 & 14.81 & 15.21 & 15.25 & 15.56 \\ \hline
Speedup & 5.63 & 5.76 & 6.69 & 7.75 & 6.03 & 6.49 & 12.85 & 18.26 & 21.63 & 27.89 \\ \hline
\textbf{ALL-GPU DE (GPU)} & 1.35 & 1.34 & 1.33 & 1.38 & 1.42 & 1.50 & 1.77 & 1.99 & 2.27 & 2.57 \\ \hline
Speedup & 59.82 & 60.44 & 71.11 & 81.67 & 61.10 & 62.86 & 107.78 & 139.28 & 145.61 & 168.69 \\ \hline
\textbf{ALL-GPU iDE (GPU)} & 0.63 & 0.61 & 0.60 & 0.66 & 0.70 & 0.78 & 1.05 & 1.28 & 1.55 & 1.85 \\ \hline
Speedup & 128.68 & 132.58 & 156.38 & 170.64 & 123.94 & 121.12 & 181.62 & 217.49 & 212.78 & 234.87 \\ \hline
\end{tabular}%
\end{table*}

\begin{figure}[!t]
    \centering
    \includegraphics[width=\textwidth]{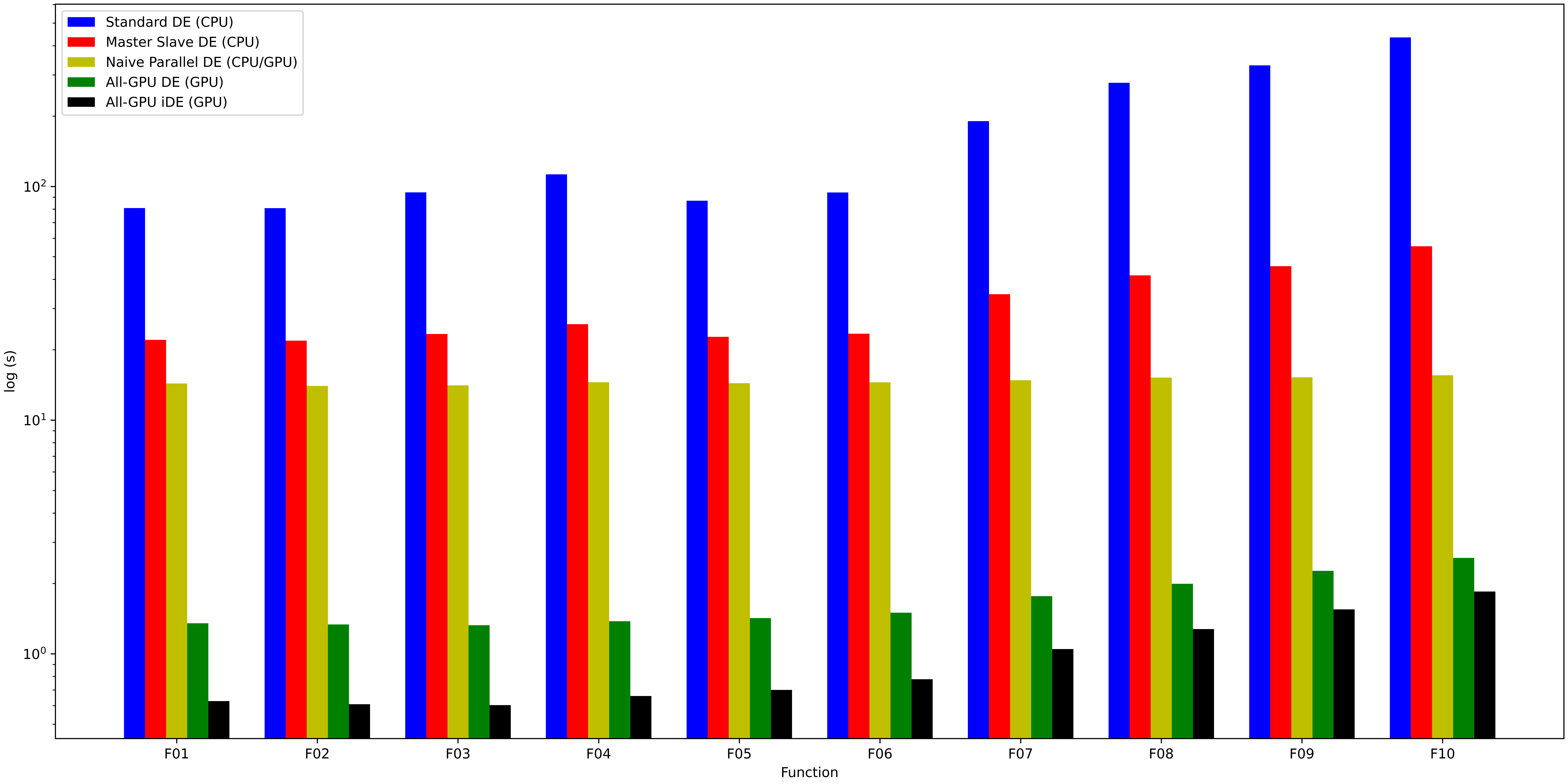}
    \caption{A logarithmic bar plot showing the wall-clock time in seconds for five different implementations of differential evolution on the proposed benchmark. The All-GPU iDE achieved a speedup ranging from 121x to 234.9x compared to the standard DE algorithm.}
    \label{fig:benchmark_compare}
\end{figure}

The experiment used the \textit{rand/1/bin} DE algorithm, with a fixed population size of $N_P=250$, $C_R=0.3$ and $F=0.5$ for all configurations on the 100-dimensional benchmark functions. The standard DE algorithm is not capable of solving any of the benchmark functions, so all configurations are executing the entire budget of 10,000,000 function evaluations and provide similar optimisation results given the differing random number generators. The average time taken for 30 runs of each algorithm on each function is presented in table \ref{tab:benchmark_compare} and shown in Fig. \ref{fig:benchmark_compare}. 

The complexity of the objective function greatly impacts the overall run-time of the algorithm, which is especially apparent with the composition functions F08, F09, and F10. These functions are comprised of 3, 4, or 5 basic functions which are each evaluated and combined into a single objective value, therefore the computation time for the standard single threaded approach is around 3, 4, or 5 times longer than any single basic function. The CPU based master slave model used a thread pool of 20 threads and provided a wall-clock speedup of between 3.7x to 7.8x, where the largest speedup is found when optimising the composition functions. 

The overhead of passing data to the thread pool for evaluation is apparent, with the simpler objective functions obtaining smaller computational speedups. This trend continues when incorporating the GPU into the DE algorithm, where the Naive parallel GPU model provides a speedup of between 5.6x and 28x against the single-threaded approach. This experiment shows the benefit of moving the evaluation of the benchmark functions to the GPU, where a larger speedup is obtained when compared to the CPU based master slave approach. However, a clear bottleneck arises from the need to transfer the population over PCI-E to the device, evaluate the fitness values, and then transfer them back to the host during each iteration.

With a population size of 250, and a computational budget of 10,000,000 function evaluations, this transfer of data is occurring 40,000 times. To address this, the All-GPU model eliminates the PCI-E bottleneck by executing entirely on the GPU, resulting in a significant improvement in wall-clock time. Specifically, the All-GPU model achieved a speedup ranging from 59.8x to 168.7x compared to the single-threaded approach. When comparing the All-GPU model with the multi-threaded CPU Master Slave approach, a speedup of 16.3x to 21.6x was obtained. Additionally, when comparing the All-GPU parallel model to the Naive parallel model, a speedup of 6x to 10.6x was achieved.

Finally, when consolidating logically related kernels and utilising CUDA streams, the All-GPU iDE provided a speedup of 121x to 234.9x against the single-threaded CPU DE. Compared against the CPU based Master Slave a speedup of 29.4x to 39x was achieved, and a speedup of 8.4x to 23.4x against the Naive parallel model. This implementation is 1.4x to 2.2x faster than the All-GPU DE, showing the advantages of kernel consolidation, concurrency using streams, and reduced global memory accesses. 

The experiment demonstrated the significant wall-clock speedups achieved by developing GPU-based DE algorithms. While the kernel-based approach utilised in the All-GPU DE effectively eliminated the PCI-E bottleneck, it was evident that further speed improvements could be attained through careful design of the CUDA kernels as found in the All-GPU iDE model.

\begin{figure}[!t]
    \centering
    \includegraphics[width=88mm]{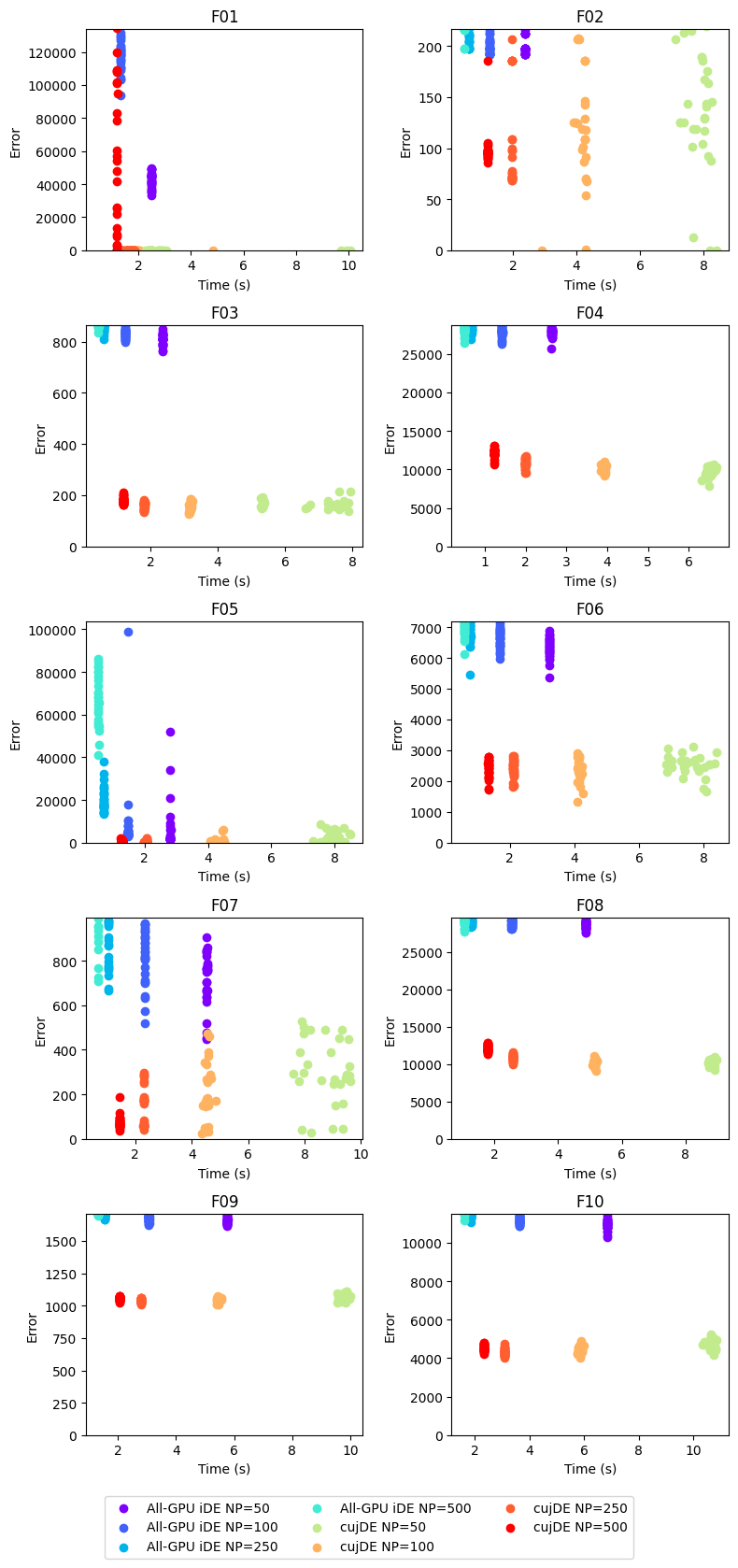}
    \caption{Trials of two algorithms All-GPU iDE and cujDE with population sizes 50, 100, 250, 500, run for 30 trials each on each of the proposed benchmark functions.}
    \label{fig:ide_vs_jde}
\end{figure}

\begin{table*}[!t]
\caption{Ranking of All-GPU iDE and cujDE with Population Sizes 50, 100, 250, 500, on the Proposed Benchmark and the Final Ranking is Included in the Last Column.}
\label{tab:de_jde_rank}
\centering
\begin{tabular}{cllllllllllll}
\hline
\multicolumn{1}{l}{} & \textbf{$N_P$} & \textbf{F01} & \textbf{F02} & \textbf{F03} & \textbf{F04} & \textbf{F05} & \textbf{F06} & \textbf{F07} & \textbf{F08} & \textbf{F09} & \textbf{F10} & \textbf{Rank} \\ \hline
\multirow{4}{*}{\textbf{All-GPU iDE}} & 50 & 4 & 6 & 5 & 5 & 5 & 5 & 5 & 5 & 5 & 5 & 5 \\ \cline{2-13} 
 & 100 & 6 & 5 & 6 & 6 & 6 & 6 & 6 & 6 & 6 & 6 & 6 \\ \cline{2-13} 
 & 250 & 7 & 7 & 7 & 7 & 7 & 7 & 7 & 7 & 7 & 7 & 7 \\ \cline{2-13} 
 & 500 & 8 & 8 & 8 & 8 & 8 & 8 & 8 & 8 & 8 & 8 & 8 \\ \hline
\multirow{4}{*}{\textbf{cujDE}} & 50 & 3 & 4 & 3 & 1 & 4 & 4 & 4 & 2 & 4 & 4 & 3 \\ \cline{2-13} 
 & 100 & 2 & 3 & 1 & 2 & 3 & 2 & 3 & 1 & 2 & 2 & 2 \\ \cline{2-13} 
 & 250 & 1 & 2 & 2 & 3 & 1 & 3 & 2 & 3 & 1 & 1 & 1 \\ \cline{2-13} 
 & 500 & 5 & 1 & 4 & 4 & 2 & 1 & 1 & 4 & 3 & 3 & 4 \\ \hline
\end{tabular}%
\end{table*}

\subsection{Case study 2}
This experiment presents an example of using the evaluation criteria and comparing the optimisation accuracy between differing population sizes of the All-GPU iDE and the open source cujDE \cite{ref:2020:mateus} on the proposed benchmark. The ranking of results are shown Table \ref{tab:de_jde_rank} and a comparison between error and wall-clock time is presented in Fig. \ref{fig:ide_vs_jde}. As expected, the cujDE algorithm outperforms the All-GPU in all benchmark functions due to its ability to dynamically adapt the $F$ and $C_R$ values during run-time. This self-adaptive feature enhances the optimisation algorithm, resulting in superior performance.

The optimal population size for cujDE varied according to the objective function. For F04, the best performance was achieved with a population size of $N_P=50$, while F03 and F08 showed superior results with $N_P=100$. F01, F05, F09, and F10 performed best with a population size of $N_P=250$. Lastly, F02, F06, and F07 achieved the best results with a population size of $N_P=500$. Overall, a population size $N_P=250$ had the best overall performance on the benchmark. 

Notably, cujDE managed to find solutions for some trials in F01 and F02. To rank these successful trials, the wall-clock time was used as the criteria. This experiment also reiterated the findings of previous works that increasing the population size, although giving a better wall-clock speedup with respect to function evaluations, does not necessarily improve convergence properties of an algorithm, and can hinder the search depending on the objective function.

% -----------------------------
% ---------- SECTION ----------
% -----------------------------
\section{Conclusion}
\label{sec:conclusion}
This paper provides an overview of select works that focus on the mapping of DE onto the GPU architecture. In the first few years (2010-2012) a lot of progress was made in mapping DE to suit and exploit the GPU architecture. Interestingly, although the most recent works have begun exploring GPU based implementations of self-adaptive DE algorithms, they have returned to a kernel for each operation in the GPU-based DE algorithm. Most of the implementations have compared the correctness and performance of their algorithm to a sequential CPU implementation, with little to no comparison between different GPU-based algorithms. 

In addition, this work also introduced a GPU-based benchmark built from objective functions taken from CEC competitions. The results clearly demonstrated an immense speedup can be achieved by executing both the algorithm and objective functions on the GPU, surpassing the performance of both single-threaded and multi-threaded CPU implementations. This benchmark can now set a baseline for future GPU based DE algorithm development and competition. Furthermore, it can be extended for application in other EAs, including genetic algorithms and particle swarm optimisation. 

As found in case study 1, the All-GPU iDE, which incorporates consolidated kernels and utilises CUDA streams, demonstrated improved wall-clock performance compared to using a separate kernel for each operation. This suggests that when implementing other variations of DE, adopting this approach could be beneficial. In case study 2, cujDE was shown to have superior convergence properties compared to the All-GPU iDE. However, it is important to note that neither algorithm achieved satisfactory solutions for majority of the benchmark problems, indicating the potential for further development of GPU-based DE algorithms.

Future work should explore consolidating kernels further to facilitate shared memory reuse, minimise global memory accesses, and utilise streams for concurrent processing. Additionally, exploring newer and more efficient variants of DE, such as history-based adaptive DE\cite{ref:2009:zhang}, holds great potential for improving performance and convergence properties. 

Additionally, the multi-population parallel model, which has proven successful in GPU-based genetic algorithms\cite{ref:2022:janssen}, could be extended to GPU-based DE algorithms as well. These avenues hold promising potential for enhancing the performance of GPU-based DE algorithms.

\bmsection*{Conflict of interest}

The authors declare no potential conflict of interests.

\bibliography{wileyNJD-AMA}

\end{document}